\documentclass[conference]{IEEEtran}
\IEEEoverridecommandlockouts
\usepackage{cite}
\usepackage{amsmath,amssymb,amsfonts}
\usepackage{algorithmic}
\usepackage{graphicx}
\usepackage{textcomp}
\usepackage{booktabs} 
\usepackage{adjustbox} 
\usepackage{caption} 
\usepackage{graphicx}
\usepackage{bbm} 

\usepackage{textcomp}
\usepackage[hidelinks]{hyperref} 
\usepackage{subcaption}
\usepackage{array} 
\usepackage{xcolor}
\def\BibTeX{{\rm B\kern-.05em{\sc i\kern-.025em b}\kern-.08em
    T\kern-.1667em\lower.7ex\hbox{E}\kern-.125emX}}
\begin{document}

\title{Pruning for Generalization: A Transfer-Oriented Spatiotemporal Graph Framework}

\author{\IEEEauthorblockN{1\textsuperscript{st} Zihao Jing}
\IEEEauthorblockA{\textit{Department of Computer Science} \\
\textit{Western University}\\
London, Canada \\
zjing29@uwo.ca}
\and
\IEEEauthorblockN{2\textsuperscript{nd} Yuxi Long}
\IEEEauthorblockA{\textit{Department of Computer Science} \\
\textit{Western University}\\
London, Canada \\
ylong66@uwo.ca}
\and
\IEEEauthorblockN{3\textsuperscript{rd} Ganlin Feng}
\IEEEauthorblockA{\textit{Department of Computer Science} \\
\textit{Western University}\\
London, Canada \\
gfeng23@uwo.ca}
}

\maketitle

\begin{abstract}
Multivariate time series forecasting in graph-structured domains is critical for real-world applications, yet existing spatiotemporal models often suffer from performance degradation under data scarcity and cross-domain shifts. We address these challenges through the lens of structure-aware context selection. We propose \textbf{TL-GPSTGN}, a transfer-oriented spatiotemporal framework that enhances sample efficiency and out-of-distribution generalization by selectively pruning non-optimized graph context. Specifically, our method employs information-theoretic and correlation-based criteria to extract structurally informative subgraphs and features, resulting in a compact, semantically grounded representation. This optimized context is subsequently integrated into a spatiotemporal convolutional architecture to capture complex multivariate dynamics. Evaluations on large-scale traffic benchmarks demonstrate that TL-GPSTGN consistently outperforms baselines in low-data transfer scenarios. Our findings suggest that explicit context pruning serves as a powerful inductive bias for improving the robustness of graph-based forecasting models. \underline{\href{https://anonymous.4open.science/r/GP-TLSTGCN-22C1/README.md }{Code Link}}.
\end{abstract}


\section{Introduction}
Accurate traffic forecasting is fundamental to intelligent transportation systems, enabling routing, signal control, and congestion mitigation. Yet traffic exhibits both periodic structure (e.g., commuting cycles) and non-stationary disruptions (e.g., incidents and weather), making prediction difficult. Recent deep learning methods—especially Spatiotemporal Graph Convolutional Networks (STGCNs) \cite{b8}—perform well by jointly modeling spatial dependencies on road graphs and temporal dynamics in multivariate sensor streams. However, strong performance typically relies on dense, high-quality historical data, which is often missing in newly deployed or underdeveloped networks \cite{b6}; moreover, STGCNs can overfit region-specific topology and patterns, motivating transfer learning (TL) from data-rich source regions to data-sparse targets.

A core bottleneck for TL is the fidelity of the graph context presented to the model. Traffic graphs differ across cities in topology and sensor placement, and may include weakly correlated edges, redundant structure, and boundary nodes whose signals are confounded by external regions not represented in the modeled graph. Treating such graphs as uniformly reliable injects noise into message passing, increases representation variance, and amplifies cross-domain shift, limiting out-of-distribution performance; existing TL approaches (e.g., parameter sharing, domain adaptation, distillation) largely ignore this input-level unreliability. We therefore argue for structure-aware context selection before spatiotemporal modeling: graph pruning can act as an inductive bias that suppresses low-support boundary nodes and unreliable edges, yielding a cleaner subgraph on which spatiotemporal filters learn more invariant features and adapt with fewer target labels.

In this paper, we propose \textbf{TL-GPSTGN}, a transfer-oriented spatiotemporal framework that utilizes graph pruning for cross-domain traffic forecasting. We introduce an entropy--correlation dual-criteria selector: entropy captures node-level traffic variability, while correlation measures temporally stable inter-node dependencies. Their combination yields a compact subgraph that preserves salient intra-region connectivity and suppresses boundary-induced noise. The pruned context is modeled by an STGCN backbone \cite{b8} and transferred via source pretraining followed by target fine-tuning. Experiments on large-scale benchmarks show consistent improvements in low-data transfer, indicating stronger robustness and sample efficiency across heterogeneous cities.

\paragraph{Contributions.}
(1) We show that uncurated graph context and boundary noise limit STGCN transferability. (2) We propose an entropy--correlation criterion to retain informative dependencies while pruning outer-layer nodes. (3) We integrate the pruner into a source-to-target TL pipeline and demonstrate consistent gains in data-sparse target domains.
.

\section{Background and Related Work}

\subsection{Spatiotemporal Forecasting Paradigms}
Traffic forecasting has progressed from classical time-series models (e.g., ARIMA and Kalman filters) \cite{b7,b12} to deep spatiotemporal architectures that capture nonlinear dynamics at scale. In urban settings, purely temporal predictors are often insufficient because they ignore road-network topology. STGCNs \cite{b8} address this by coupling graph convolution over an adjacency matrix $A\in\mathbb{R}^{n\times n}$ with temporal modules (e.g., 1D convolutions or attention). While effective in-domain, many STGCNs learn topology-specific propagation patterns, which can degrade under cross-domain shifts where $G_{\text{target}}$ differs from $G_{\text{source}}$ \cite{b6}.

\subsection{Problem Formalization}
We model a traffic network as $\mathcal{G}=(\mathcal{V},\mathcal{E},A)$ with $|\mathcal{V}|=n$ sensors and weighted adjacency $A$. Let $X_t\in\mathbb{R}^{n}$ be the traffic state at time $t$. Given a history window of length $H$, the task is to predict a horizon of length $T$:
\begin{equation}
\hat{X}_{t+1:t+T}=F\!\left(X_{t-H+1:t};\,\mathcal{G},\Theta\right),
\end{equation}
where $\Theta$ are model parameters. In transfer learning, we aim to minimize target-domain error on $\mathcal{D}_T$ with limited labels by leveraging a data-rich source domain $\mathcal{D}_S$.

\subsection{Limitations of Current Literature}
Prior work improves robustness and transfer via richer data sources \cite{rae2024gps,luo2024ad_hoc}, probabilistic modeling \cite{visconti2024bayesian}, or domain-oriented objectives \cite{yoosuf2024perceptual,manibardo2024forecasting}. However, many approaches treat the full adjacency matrix $A$ as fixed input during migration, leaving structural redundancy and boundary-driven noise unaddressed. This can inject irrelevant inductive bias in the target domain and reduce sample efficiency. In contrast, we focus on contextual distillation: selecting a compact, informative subgraph to stabilize spatiotemporal representation learning across heterogeneous urban topologies.

\section{Methodology}
We summarize the TL-GPSTGN pipeline and then formalize the graph pruning novelty, including the Information Entropy Analyzer (IEA) and the dual-criteria selection rule.
\begin{figure}[htbp]
\centerline{\includegraphics[width=\linewidth]{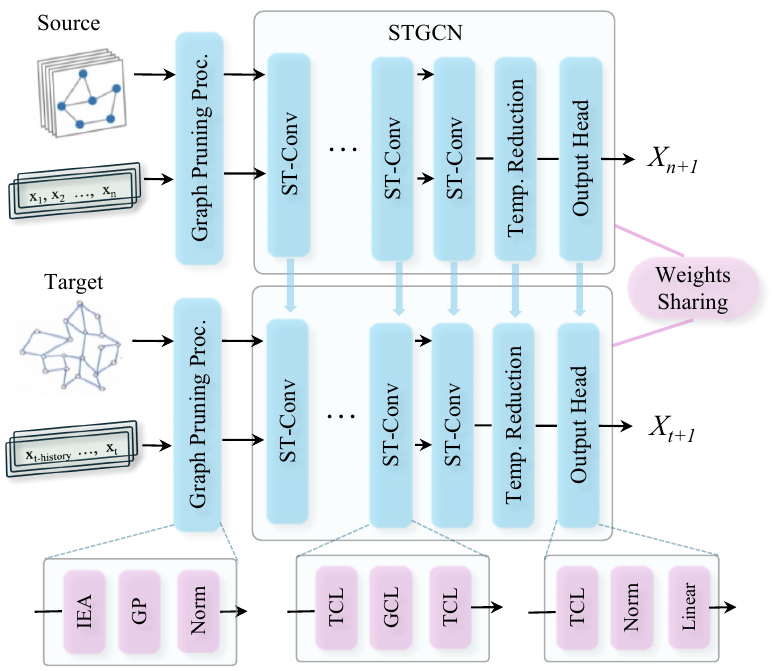}}
\caption{Overview of TL-GPSTGN. The graph pruning processor (GPP) comprises an Information Entropy Analyzer (IEA), graph pruning (GP), and normalization. Each spatiotemporal convolution (ST-Conv) block follows a TCL--GCL--TCL design, i.e., two temporal convolution layers (TCL) with a graph convolution layer (GCL) in between.}
\label{fig:structure}
\end{figure}
\subsection{Model Architecture Overview}
To enable traffic forecasting with limited target-domain labels, we propose a transfer-learning spatiotemporal graph convolutional network, \textbf{TL-GPSTGN}, consisting of: (i) a \emph{Graph Pruning Processor} (GPP), (ii) an \emph{STGCN} backbone, and (iii) a \emph{Reductor}. GPP performs entropy-correlation-based edge scoring followed by outer-layer pruning and normalization. The STGCN integrates spatial graph convolution with temporal convolution to learn on non-Euclidean road-network signals \cite{b8}. The Reductor inverts the normalization to output predictions in the original traffic-flow space. Fig.~\ref{fig:structure} illustrates the overall architecture.

Training follows two stages. \textbf{Source pretraining:} we apply GPP on a data-rich source network and train the STGCN to obtain pretrained parameters. \textbf{Target adaptation:} we apply the same GPP on the target network and fine-tune the STGCN from the pretrained initialization using a small labeled target set.

\subsection{STGCN Module}
STGCN is composed of stacked spatiotemporal convolution blocks (ST-Conv) in a \texttt{TCL--GCL--TCL} ``hamburger'' pattern \cite{b8}. The graph convolutional layer (GCL) aggregates neighborhood information via the (possibly pruned) adjacency matrix, while the temporal convolutional layer (TCL) applies 1D convolution over time to capture temporal dynamics. The final prediction head uses an additional TCL and linear layers to generate $X_{\mathrm{pred}}$.

\subsection{Graph Pruning Processor}
\label{sec:gpp}
\textbf{Motivation (core novelty).} Real road networks are open systems: boundary sensors are strongly influenced by external regions that are \emph{not} represented in the modeled graph. Such nodes tend to have weak and noisy intra-graph connectivity, degrading both in-domain learning and cross-domain transfer. Our pruning explicitly removes these \emph{outer-layer} nodes (low-degree after informative edge selection), yielding a compact subgraph that better reflects internal dynamics and transfers more robustly.

\paragraph{Notation.}
Let $G=(V,E)$ be a road graph with $|V|=N$ sensors. Let $A\in\{0,1\}^{N\times N}$ be the raw adjacency and $x_i\in\mathbb{R}^{T}$ be the historical traffic series at node $i$ over $T$ time steps (e.g., speed/flow). Denote the node-feature matrix by $X=[x_1,\dots,x_N]^\top\in\mathbb{R}^{N\times T}$.

\subsubsection{Information Entropy Analyzer (IEA)}
IEA quantifies node-wise uncertainty/variability from observed traffic, then combines it with pairwise correlation to score which edges are informative for intra-graph propagation.

\paragraph{Node entropy.}
We discretize $x_i$ into $B$ bins to obtain an empirical distribution
\begin{equation}
p_{i}(b) \;=\; \frac{1}{T}\sum_{t=1}^{T}\mathbbm{1}\!\left[x_{i,t}\in \mathcal{I}_b\right],\quad b=1,\dots,B,
\end{equation}
and define Shannon entropy
\begin{equation}
H_i \;=\; -\sum_{b=1}^{B} p_i(b)\log\big(p_i(b)+\epsilon\big),
\end{equation}
where $\epsilon>0$ avoids $\log 0$. Intuitively, larger $H_i$ indicates richer, less trivial dynamics (higher information content) at node $i$.

\paragraph{Correlation-based criterion.}
For each candidate edge $(i,j)$ with $A_{ij}=1$, we compute the (absolute) Pearson correlation
\begin{equation}
r_{ij} \;=\; \left|\frac{\sum_{t=1}^{T}(x_{i,t}-\bar{x}_i)(x_{j,t}-\bar{x}_j)}
{\sqrt{\sum_{t=1}^{T}(x_{i,t}-\bar{x}_i)^2}\sqrt{\sum_{t=1}^{T}(x_{j,t}-\bar{x}_j)^2}+\epsilon}\right|.
\end{equation}
We then form an entropy-modulated edge importance score
\begin{equation}
s_{ij} \;=\; A_{ij}\cdot r_{ij}\cdot \frac{H_i+H_j}{2}.
\label{eq:edge_score}
\end{equation}
This dual-criteria score prefers edges that (i) connect nodes with aligned temporal behavior (high $r_{ij}$) and (ii) involve informative nodes (high $H_i,H_j$), suppressing weak/noisy connections.

\subsubsection{Outer-Layer Graph Pruning (GP)}
Using the scores $\{s_{ij}\}$, we construct a pruned adjacency by thresholding (or top-$k$ per node):
\begin{equation}
\widetilde{A}_{ij} \;=\; \mathbbm{1}\!\left[s_{ij}\ge \tau\right],
\end{equation}
and define the pruned degree $\widetilde{d}_i=\sum_j \widetilde{A}_{ij}$. We identify boundary/outer-layer nodes as those with small intra-graph support:
\begin{equation}
V_{\mathrm{out}} \;=\; \{\, i\in V \mid \widetilde{d}_i \le d_{\min}\,\}.
\end{equation}
We remove $V_{\mathrm{out}}$ (and incident edges) and optionally iterate this procedure for $L$ layers to peel multiple outer rings. The final kept node set $V_{\mathrm{keep}}$ induces the compact subgraph used by the downstream STGCN. Compared with generic sparsification, this explicitly targets externally-driven boundary sensors, reducing noise and improving transferability \cite{b9}.

\subsubsection{Normalization and Reductor}
GPP normalizes the retained node signals (e.g., z-score per sensor) before STGCN training. The Reductor inverts this normalization to map predictions back to the original traffic scale.

\subsection{Transfer Learning}
We first train TL-GPSTGN on a source network with abundant history to learn transferable spatiotemporal filters, then fine-tune on a target network with limited labels. Crucially, applying the same GPP in both domains yields a more comparable (less boundary-noisy) internal subgraph, stabilizing migration and improving sample efficiency in target adaptation \cite{b10,b11}.

\begin{table}[t]
\centering
\caption{Results of single dataset evaluation (15 minutes)}
\label{tab:single_dataset_results_15}
\sc
\renewcommand{\arraystretch}{1.0}
\begin{adjustbox}{max width=0.49\textwidth}
\begin{tabular}{@{}lccccccccc@{}}
\toprule
Models & \multicolumn{3}{c}{metr-la} & \multicolumn{3}{c}{pems-bay} & \multicolumn{3}{c}{pemsd7} \\ 
\cmidrule(lr){2-4} \cmidrule(lr){5-7} \cmidrule(lr){8-10}
 & MAE & MAPE & RMSE & MAE & MAPE & RMSE & MAE & MAPE & RMSE \\ 
\midrule
HA & 4.16 & 13.0 & 7.80 & 2.88 & 6.80 & 5.59 & 4.01 & 10.61 & 4.55 \\ 
ARIMA & 3.99 & 9.6 & 8.21 & \textbf{1.62} & \textbf{3.50} & \textbf{3.30} & 5.55 & 12.92 & 9.00 \\ 
FNN & 3.99 & 9.9 & 7.94 & 2.20 & 5.19 & 4.42 & 2.74 & 6.38 & 4.75 \\ 
FC-LSTM & 3.44 & 9.6 & \textbf{6.30} & 2.05 & 4.8 & 4.19 & 3.57 & 8.60 & 6.20 \\ 
STGCN & 3.40 & 6.71 & 6.56 & 2.50 & 4.03 & 5.11 & \textbf{2.38} & \textbf{4.16} & \textbf{4.29} \\ 
TL-GPSTGN & \textbf{3.37} & \textbf{6.58} & 6.62 & 2.92 & 4.77 & 5.79 & 2.47 & 4.34 & 4.30 \\ 
\bottomrule
\end{tabular}
\end{adjustbox}
\end{table}

\begin{table}[t]
\centering
\caption{Results of single dataset evaluation (30 minutes)}
\sc
\label{tab:single_dataset_results_30}
\renewcommand{\arraystretch}{1.0}
\begin{adjustbox}{max width=0.49\textwidth}
\begin{tabular}{@{}lccccccccc@{}}
\toprule
Models & \multicolumn{3}{c}{metr-la} & \multicolumn{3}{c}{pems-bay} & \multicolumn{3}{c}{pemsd7} \\ 
\cmidrule(lr){2-4} \cmidrule(lr){5-7} \cmidrule(lr){8-10}
 & MAE & MAPE & RMSE & MAE & MAPE & RMSE & MAE & MAPE & RMSE \\ 
\midrule
HA & 4.16 & 13.0 & 7.80 & 2.88 & 6.80 & 5.59 & 4.01 & 10.61 & 6.67 \\ 
ARIMA & 5.15 & 12.7 & 10.45 & 2.33 & 8.30 & 4.76 & 5.86 & 13.94 & 9.13 \\ 
FNN & 4.23 & 12.9 & 8.17 & 2.30 & 5.43 & 4.63 & 4.02 & 9.72 & 6.98 \\ 
FC-LSTM & \textbf{3.77} & 10.9 & \textbf{7.23} & \textbf{2.20} & \textbf{5.2} & \textbf{4.55} & 3.94 & 9.55 & 7.03 \\ 
STGCN & 4.41 & \textbf{8.71} & 9.17 & 3.26 & 5.26 & 6.59 & \textbf{3.05} & \textbf{5.27} & \textbf{5.22} \\ 
TL-GPSTGN & 4.45 & \textbf{8.70} & 9.34 & 3.51 & 5.76 & 6.67 & 3.18 & 5.53 & 5.72 \\ 
\bottomrule
\end{tabular}
\end{adjustbox}
\end{table}

\section{Experiment}
We evaluate TL-GPSTGN for (i) single-dataset forecasting accuracy and (ii) cross-dataset transferability. Since our core contribution targets \emph{migration robustness}, we emphasize that TL-GPSTGN is not always the best on single-dataset forecasting, but remains competitive while consistently improving transfer performance under limited target supervision. We first describe datasets, preprocessing, and baselines, then report results on three benchmarks and two transfer settings.

\subsection{Dataset and Preprocessing}
\textbf{METR-LA} (Los Angeles) contains traffic flow from 207 inductive loop detectors, sampled over 2012--2013 (one year), provided by the Los Angeles Department of Transportation.

\textbf{PEMS-BAY} (California Bay Area) is a larger-scale dataset maintained by the ITS Lab at UC Berkeley, collected since 2007, with $>$300 detectors deployed on major roadways (freeways, city streets, bridges) and 1-minute sampling.

\textbf{PEMSD7} is a large-scale real-time PeMS dataset (ITS Lab, UC Berkeley) with 200 detectors and 5-minute sampling.

We apply \textbf{standardization} (zero mean, unit variance) to improve feature comparability and training stability. For reproducibility, normalization statistics (mean, standard deviation) are computed on the \emph{training split only} and reused unchanged for validation and test splits.

\subsection{Evaluation Metrics and Baseline}
We report \textbf{MAE}, \textbf{RMSE}, and \textbf{MAPE} for traffic forecasting evaluation \cite{b11}. Baselines include: \textbf{HA}, \textbf{ARIMA} \cite{b12}, \textbf{FNN} \cite{b13}, and \textbf{FC-LSTM} \cite{b14}. We additionally compare against \textbf{STGCN} \cite{b8}, which is a strong single-dataset baseline and the transfer baseline in migration experiments.

\subsection{Experiments and Results}
\textbf{Setup.} All experiments run on two RTX4090 GPUs (24\,GB) and complete in $\sim$72 hours. Data are split into train/val/test with a ratio of 7:1.5:1.5. Early stopping is used with a maximum of 200 epochs. The history length is 12; prediction horizons are 15 and 30 minutes. For PEMS-BAY (1-minute sampling), the input length for the 30-minute task is set to 24. We fix the random seed to 42. STGCN uses 2 ST-Conv modules; SCL and TCL kernel sizes are 3. Batch size is 32; learning rate is 0.001; L2 regularization with weight decay 0.0005; optimizer is Adam.

\begin{figure}[t]
\centering
\includegraphics[width=\linewidth]{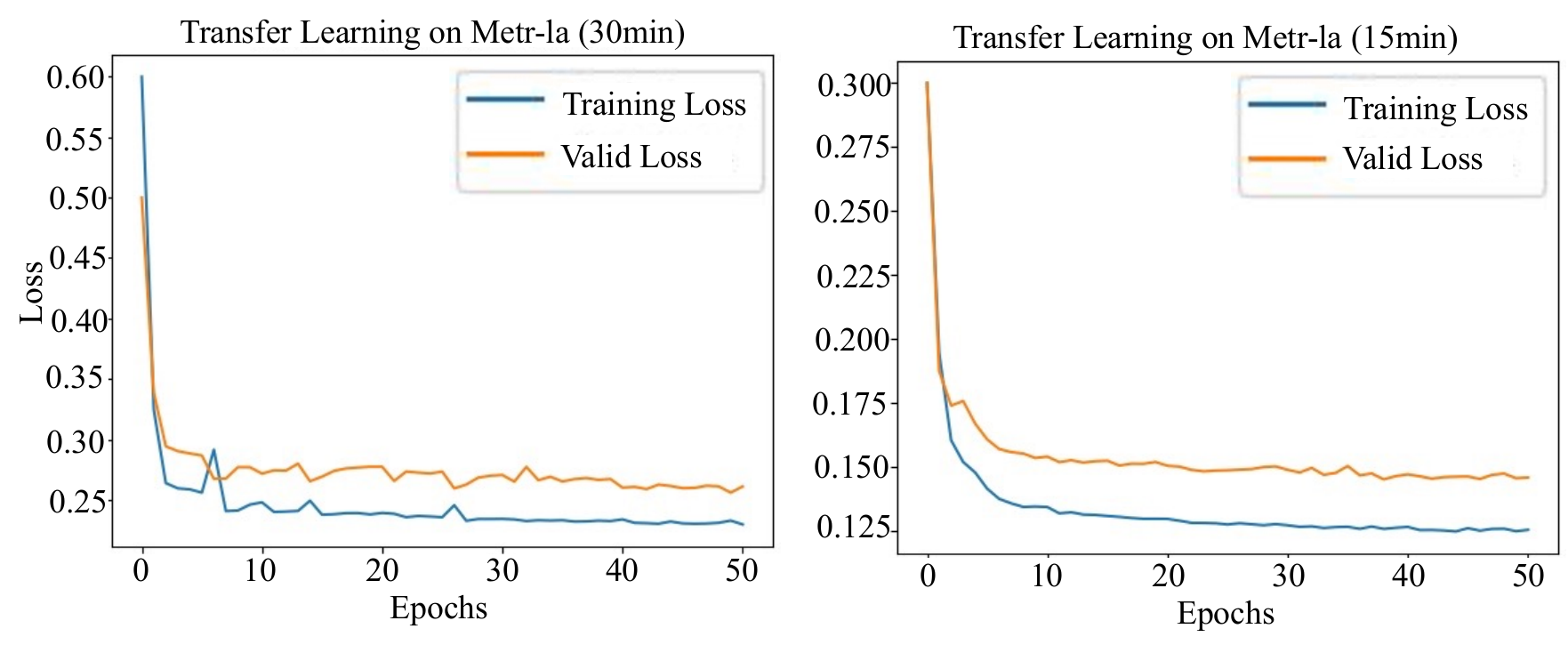}
\vskip -10 pt
\small \caption{Training and test loss of TL-GPSTGN on metr-la. Left: 15-minute horizon. Right: 30-minute horizon. Both curves exhibit rapid early descent followed by stabilization, indicating convergence and stable generalization.}
\label{fig:loss_metr}
\vskip -10 pt
\end{figure}

\textbf{Single-dataset evaluation.} We evaluate STGCN and TL-GPSTGN on metr-la, pems-bay, and pemsd7-m at 15/30-minute horizons, compute MAE/MAPE/RMSE, and compare with HA/ARIMA/FNN/FC-LSTM under matched conditions. Tables~\ref{tab:single_dataset_results_15}--\ref{tab:single_dataset_results_30} summarize quantitative results; Fig.~\ref{fig:loss_metr} reports representative convergence on metr-la.

Tables~\ref{tab:single_dataset_results_15}--\ref{tab:single_dataset_results_30} show that STGCN and TL-GPSTGN outperform HA/ARIMA/FNN/FC-LSTM across datasets and horizons. TL-GPSTGN is competitive but not uniformly better than STGCN in-domain (e.g., worse on pems-bay/pemsd7 at 15 minutes and across datasets at 30 minutes), suggesting that pruning mainly preserves predictive structure while reducing graph complexity and noise, rather than directly optimizing single-dataset error. Fig.~\ref{fig:loss_metr} reports representative convergence on metr-la.

\begin{figure}[t]
\centering
 \includegraphics[width=\linewidth]{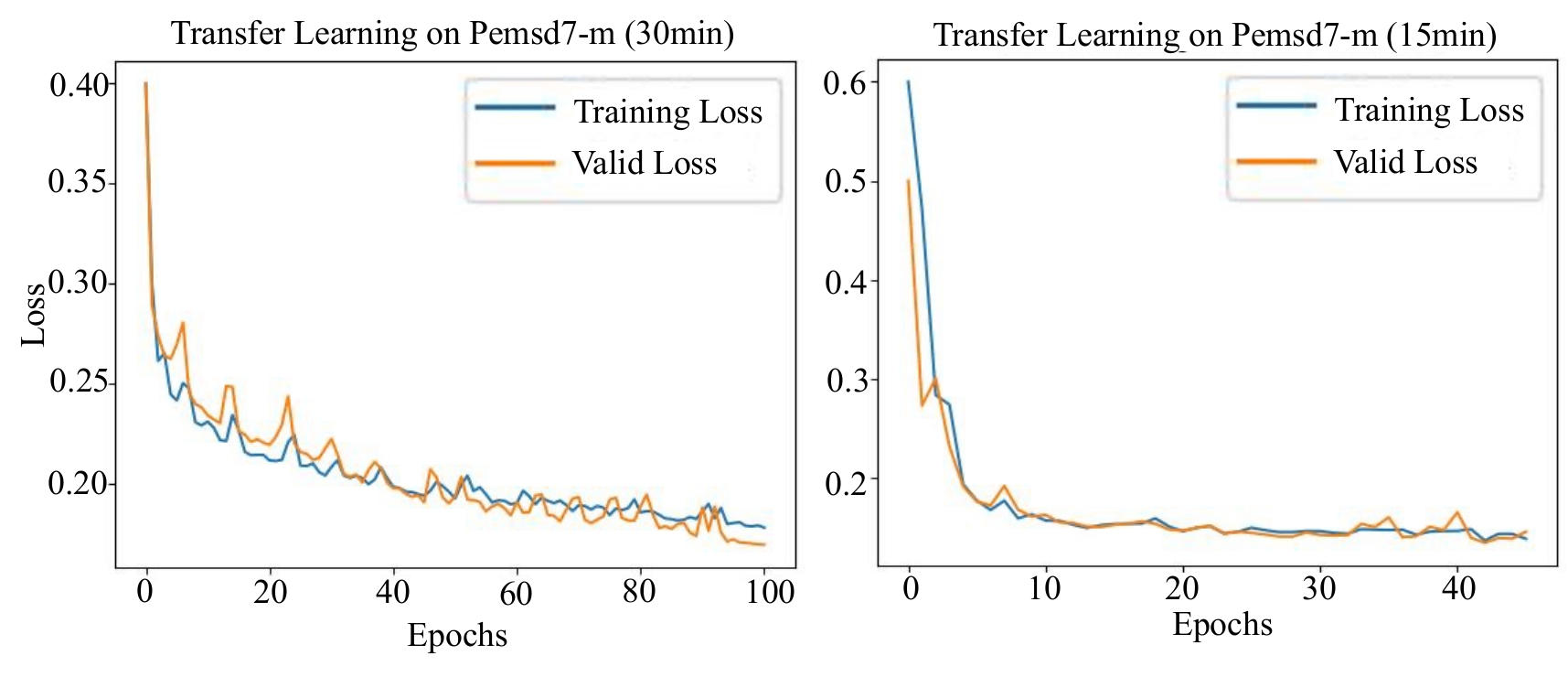}
\vskip -10 pt
\caption{Transfer-learning loss on pemsd7-m. Left: 15-minute horizon. Right: 30-minute horizon. Curves decrease steadily and converge, indicating effective adaptation to the target.}
\label{fig:loss_tf_pemsd7}
\end{figure}

\textbf{Migration evaluation.} We compare transfer performance of STGCN and TL-GPSTGN by pretraining both on metr-la, then adapting to pemsd7-m and pems-bay using limited target data. For pemsd7-m, we evaluate both 15- and 30-minute horizons; for pems-bay, we evaluate the 15-minute horizon. We vary \textbf{Target/Source (T/S)} ratio to measure adaptation under different target-data budgets. Fig.~\ref{fig:loss_tf_pemsd7} visualizes transfer-learning loss on pemsd7-m; Tab.~\ref{tab:transfer_capability} reports MAE/MAPE/RMSE.

\begin{table}[t]
\centering
\caption{Transfer capability evaluation of different models}
\label{tab:transfer_capability}
\renewcommand{\arraystretch}{1.0}
\begin{adjustbox}{max width=0.5\textwidth}\sc
\begin{tabular}{@{}llccccccccc@{}}
\toprule
Model & T/S & \multicolumn{3}{c}{Metr-la To pemsd7-m} & \multicolumn{3}{c}{Metr-la To pems-bay} \\ 
\cmidrule(lr){3-5} \cmidrule(lr){6-8}
 & & MAE & MAPE (\%) & RMSE & MAE & MAPE (\%) & RMSE \\ 
 & & (15/30min) & (15/30min) & (15/30min) & (15min) & (15min) & (15min) \\ 
\midrule
STGCN & 5\%  & 3.34/4.80 & 5.83/8.36 & 5.65/8.37 & 4.71 & 7.73 & 9.40 \\ 
      & 10\% & 3.24/4.48 & 5.65/7.79 & 5.27/7.50 & 4.17 & 6.84 & 7.54 \\ 
      & 15\% & 2.83/4.00 & 4.93/6.96 & 4.76/6.73 & 4.29 & 7.04 & 7.18 \\ 
      & 20\% & 2.76/3.92 & 4.81/6.82 & 4.74/6.70 & 4.11 & 6.74 & 7.11 \\ 
      & 25\% & 2.74/3.63 & 4.77/6.41 & 4.67/6.47 & 4.69 & 7.70 & 7.64 \\ 
TL-GPSTGN & 5\%  & \textbf{3.22}/\textbf{4.40} & \textbf{5.52}/\textbf{7.55} & \textbf{5.51}/\textbf{7.79} & \textbf{3.57} & \textbf{5.76} & \textbf{7.41} \\ 
          & 10\% & \textbf{3.09}/\textbf{4.37} & \textbf{5.29}/\textbf{7.51} & \textbf{5.17}/\textbf{7.40} & \textbf{3.58} & \textbf{5.76} & \textbf{7.20} \\ 
          & 15\% & \textbf{2.56}/\textbf{3.50} & \textbf{4.39}/\textbf{6.01} & \textbf{4.44}/\textbf{6.12} & \textbf{3.90} & \textbf{6.27} & \textbf{6.79} \\ 
          & 20\% & \textbf{2.49}/\textbf{3.67} & \textbf{4.28}/\textbf{6.29} & \textbf{4.43}/\textbf{6.29} & \textbf{3.10} & \textbf{5.00} & \textbf{5.79} \\ 
          & 25\% & \textbf{2.56}/\textbf{3.41} & \textbf{4.38}/\textbf{5.84} & \textbf{4.43}/\textbf{6.07} & \textbf{3.07} & \textbf{4.94} & \textbf{5.80} \\ 
\bottomrule
\end{tabular}
\end{adjustbox}
\end{table}

Tab.~\ref{tab:transfer_capability} indicates monotonic improvement for both models as T/S increases. TL-GPSTGN consistently outperforms STGCN across target datasets and horizons at matched T/S, supporting that pruning distills transfer-relevant structure and improves migration under limited target data.

\textbf{Qualitative predictions.} We visualize transferred predictions at 10-minute resolution to assess pattern fidelity.

\begin{figure}[t]
\centering
\begin{subfigure}{0.24\textwidth}
    \includegraphics[width=\linewidth]{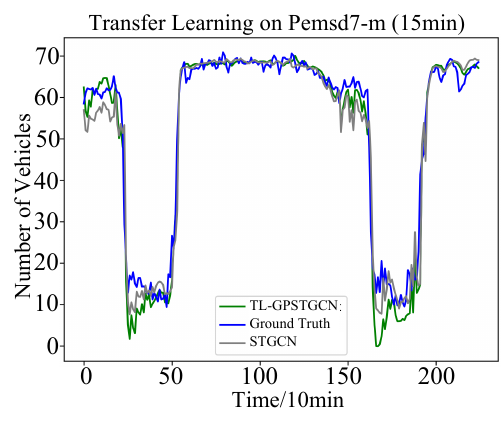}
    \label{fig:pred_tf_pemsd7_15}
\end{subfigure}
\begin{subfigure}{0.24\textwidth}
    \includegraphics[width=\linewidth]{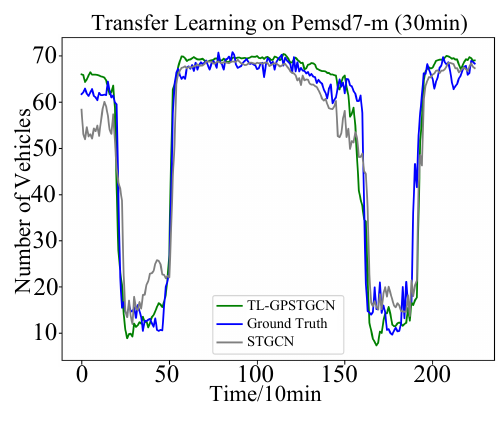}
    \label{fig:pred_tf_pemsd7_30}
\end{subfigure}
\vskip -15 pt
\caption{Partial transfer-learning predictions on pemsd7-m. Left: 15-minute horizon. Right: 30-minute horizon. TL-GPSTGN tracks ground truth more closely than STGCN, particularly around sharp transitions and extreme values.}
\label{fig:pred_tf_pemsd7}
\vskip -10 pt
\end{figure}

\begin{figure}[t]
\centering
\begin{subfigure}{0.24\textwidth}
    \includegraphics[width=\linewidth]{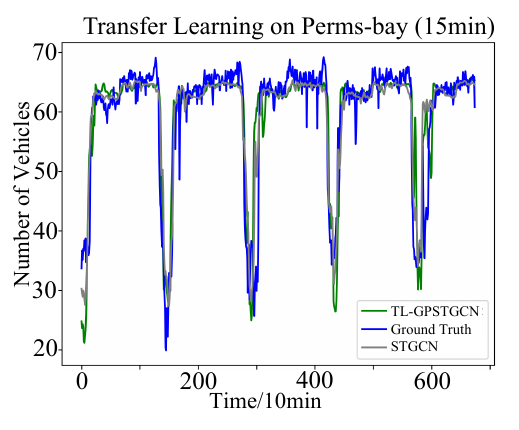}
    \label{fig:pred_tf_pemsbay_15}
\end{subfigure}
\begin{subfigure}{0.24\textwidth}
    \includegraphics[width=\linewidth]{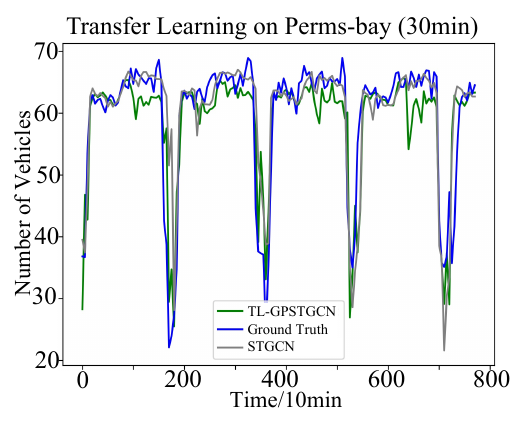}
    \label{fig:pred_tf_pemsbay_30}
\end{subfigure}
\vskip -15 pt
\caption{Partial transfer-learning predictions on pems-bay. Left: 15-minute horizon. Right: 30-minute horizon. TL-GPSTGN exhibits tighter alignment with ground truth than STGCN and adapts effectively to the new network.}
\label{fig:pred_tf_pemsbay}
\vskip -10 pt
\end{figure}

Figs.~\ref{fig:pred_tf_pemsd7}--\ref{fig:pred_tf_pemsbay} show stronger short-horizon fit (15 minutes) for both models. TL-GPSTGN better preserves peaks/troughs and sharp regime shifts under transfer, aligning with its consistent gains in Tab.~\ref{tab:transfer_capability}.

\section{Conclusion}
We proposed TL-GPSTGN, a transfer-learning STGCN framework that alleviates limited target-domain history via entropy--correlation guided graph pruning, removing low-contribution nodes/edges while preserving salient spatiotemporal dependencies. Experiments on METR-LA, PEMS-BAY, and PEMSD7 show that TL-GPSTGN consistently outperforms classical baselines (HA, ARIMA, FNN, FC-LSTM), matches STGCN in single-dataset forecasting, and achieves stronger cross-dataset migration under scarce target labels, indicating robust adaptation across heterogeneous road-network topologies. Future work will incorporate exogenous signals (e.g., weather, events, incidents), develop adaptive and data-driven pruning policies, extend to multi-modal transportation systems, and leverage self-/unsupervised pretraining to further reduce reliance on labeled target data.



\bibliographystyle{IEEEtran}
\bibliography{reference}

@article{b7,
  author  = {Nagy, A. M. and Vilmos, S.},
  title   = {Survey on traffic prediction in smart cities},
  journal = {Pervasive and Mobile Computing},
  volume  = {50},
  pages   = {148--163},
  year    = {2018}
}

@misc{b8,
  title={Spatio-temporal graph convolutional networks: A deep learning framework for traffic forecasting},
  author={Yu, Bing and Yin, Haoteng and Zhu, Zhanxing},
  journal={arXiv preprint arXiv:1709.04875},
  year={2017}
}

@article{b6,
  title={A survey on modern deep neural network for traffic prediction: Trends, methods and challenges},
  author={Tedjopurnomo, David Alexander and Bao, Zhifeng and Zheng, Baihua and Choudhury, Farhana Murtaza and Qin, Alex Kai},
  journal={IEEE Transactions on Knowledge and Data Engineering},
  volume={34},
  number={4},
  pages={1544--1561},
  year={2020},
  publisher={IEEE}
}

@article{rae2024gps,
  author  = {Raei, B. and Kinateder, M. and B{\'e}nichou, N. and Gomaa, I.},
  title   = {Are the Data Good Enough? Spatial and Temporal Modelling of Evacuee Behavior Using GPS Data},
  journal = {International Journal of Disaster Risk Reduction},
  year    = {2024}
}

@article{yoosuf2024perceptual,
  author  = {Yoosuf, S. and Baali, H. and Bouzerdoum, A.},
  title   = {Improving Perceptual Quality in Spatiotemporal Timeseries Forecasting},
  journal = {SSRN Electronic Journal},
  year    = {2024}
}

@inproceedings{luo2024ad_hoc,
  author    = {Luo, R. Q. and Liu, Y. and Wu, Y.},
  title     = {Research on Real-Time Traffic Communication Prediction Method Based on Vehicle Ad-Hoc Network Technology},
  booktitle = {Proceedings of the SPIE},
  year      = {2024}
}

@misc{visconti2024bayesian,
  author       = {Visconti, E. and Nenzi, L. and Cadonna, A.},
  title        = {Bayesian Machine Learning Meets Formal Methods: An Application to Spatio-Temporal Data},
  howpublished = {arXiv},
  eprint       = {2110.01360},
  archivePrefix= {arXiv},
  year         = {2021}
}

@misc{manibardo2024forecasting,
  author       = {Manibardo, E. L.},
  title        = {Beyond Short-Term Traffic Forecasting Models: Navigating Through Data Availability Constraints},
  howpublished = {Dialnet},
  year         = {2024}
}

@inproceedings{b9,
  title={Fast shortest-path distance queries on road networks by pruned highway labeling},
  author={Akiba, Takuya and Iwata, Yoichi and Kawarabayashi, Ken-ichi and Kawata, Yuki},
  booktitle={2014 Proceedings of the sixteenth workshop on algorithm engineering and experiments (ALENEX)},
  pages={147--154},
  year={2014},
  organization={SIAM}
}

@inproceedings{b10,
  title={Transfer learning with graph neural networks for short-term highway traffic forecasting},
  author={Mallick, Tanwi and Balaprakash, Prasanna and Rask, Eric and Macfarlane, Jane},
  booktitle={2020 25th international conference on pattern recognition (ICPR)},
  pages={10367--10374},
  year={2021},
  organization={IEEE}
}

@article{b11,
  title={Transfer learning with spatial--temporal graph convolutional network for traffic prediction},
  author={Yao, Zhixiu and Xia, Shichao and Li, Yun and Wu, Guangfu and Zuo, Linli},
  journal={IEEE Transactions on Intelligent Transportation Systems},
  volume={24},
  number={8},
  pages={8592--8605},
  year={2023},
  publisher={IEEE}
}

@article{b12,
  author  = {Kumar, S. V. and Lelitha, V.},
  title   = {Short-term traffic flow prediction using seasonal ARIMA model with limited input data},
  journal = {European Transport Research Review},
  volume  = {7},
  number  = {3},
  pages   = {1--9},
  year    = {2015}
}

@article{b13,
  title={Computer network traffic prediction: a comparison between traditional and deep learning neural networks},
  author={Oliveira, Tiago Prado and Barbar, Jamil Salem and Soares, Alexsandro Santos},
  journal={International Journal of Big Data Intelligence},
  volume={3},
  number={1},
  pages={28--37},
  year={2016},
  publisher={Inderscience Publishers (IEL)}
}

@article{b14,
  title={Human action recognition using convolutional LSTM and fully-connected LSTM with different attentions},
  author={Zhang, Zufan and Lv, Zongming and Gan, Chenquan and Zhu, Qingyi},
  journal={Neurocomputing},
  volume={410},
  pages={304--316},
  year={2020},
  publisher={Elsevier}
}

\appendix

\section{Generalization Analysis}
\label{app:gen}

This appendix provides a compact, assumption-explicit Rademacher complexity bound for the main building blocks of our model.
We use standard contraction/composition properties to propagate complexity through (i) linear convolutions, (ii) causal temporal convolutions with residual connections, and (iii) stacked blocks.
Our goal is to justify how norm control (e.g., weight decay) and parameter sharing (convolutions) constrain capacity.

\subsection{Preliminaries: empirical Rademacher complexity and a generalization bound}
Let $S=\{x_i\}_{i=1}^m$ be a sample and $\sigma_i\overset{i.i.d.}{\sim}\mathrm{Unif}\{\pm1\}$.
For a real-valued function class $\mathcal{F}$, the empirical Rademacher complexity is
\begin{equation}
\hat{\mathfrak{R}}_S(\mathcal{F})
~:=~
\mathbb{E}_{\sigma}\Big[\sup_{f\in\mathcal{F}}\frac{1}{m}\sum_{i=1}^m\sigma_i f(x_i)\Big].
\end{equation}
We will repeatedly use the following standard properties:
\begin{align}
\hat{\mathfrak{R}}_S(\mathcal{F}+\mathcal{G})
&\le \hat{\mathfrak{R}}_S(\mathcal{F})+\hat{\mathfrak{R}}_S(\mathcal{G}),
\label{eq:rad-sum}\\
\hat{\mathfrak{R}}_S(\phi\circ\mathcal{F})
&\le L_{\phi}\,\hat{\mathfrak{R}}_S(\mathcal{F})
\quad \text{if $\phi$ is $L_{\phi}$-Lipschitz.}
\label{eq:rad-contract}
\end{align}
For a loss $\ell(\cdot,y)$ that is $1$-Lipschitz in its first argument and bounded in $[0,1]$ (e.g., clipped absolute loss),
one obtains the standard uniform generalization bound: with probability at least $1-\delta$,
\begin{equation}
\begin{aligned}
\forall f\in\mathcal{F}:\quad
\mathbb{E}\,\ell\!\big(f(x),y\big)
~\le~
\frac{1}{m}\sum_{i=1}^m \ell\!\big(f(x_i),y_i\big)
~+~ 2\,\hat{\mathfrak{R}}_S(\mathcal{F}) \\
\;+\;
3\sqrt{\frac{\log(2/\delta)}{2m}}.
\end{aligned}
\label{eq:gen-bound}
\end{equation}

\subsection{A linear-layer lemma (used for convolutions)}
We will reduce each convolution to a linear map over a (vectorized) local feature representation.
Let $\psi(x)\in\mathbb{R}^d$ denote the feature vector produced by extracting and vectorizing the relevant receptive field(s).
Consider the linear class
\begin{equation}
\mathcal{F}_{\Lambda}
~:=~
\Big\{x\mapsto \langle w,\psi(x)\rangle ~:~ \|w\|_2\le \Lambda\Big\}.
\end{equation}
If $\|\psi(x_i)\|_2\le B$ for all $i$, then
\begin{equation}
\hat{\mathfrak{R}}_S(\mathcal{F}_{\Lambda})
~\le~
\frac{\Lambda B}{\sqrt{m}}.
\label{eq:linear-rad}
\end{equation}
\noindent
\textbf{Remark.}
Equation~\eqref{eq:linear-rad} is typically stated with an $\ell_2$-constraint and $\ell_2$-bounded features.
If one instead constrains $\|w\|_F$ after tensorization, the same scaling holds with the matching Frobenius feature norm.

\subsection{Rademacher complexity of a convolution layer}
A convolution layer is a linear operator with \emph{shared} parameters.
Let $W_{\mathrm{conv}}\in\mathbb{R}^{C_{\mathrm{out}}\times C_{\mathrm{in}}\times K_h\times K_w}$.
For each input $x$, define $\Phi(x)$ as the collection of all local patches (vectorized) aggregated into a single feature vector.
Then the layer output at any fixed spatial/temporal index is linear in $W_{\mathrm{conv}}$:
\begin{equation}
h(x)~=~\langle W_{\mathrm{conv}},\Phi(x)\rangle.
\end{equation}
Assume a uniform bound $\|\Phi(x_i)\|_F\le B_{\Phi}$ and a norm constraint $\|W_{\mathrm{conv}}\|_F\le \Lambda_{\mathrm{conv}}$.
Applying~\eqref{eq:linear-rad} yields
\begin{equation}
\hat{\mathfrak{R}}_S(\mathrm{Conv})
~\le~
\frac{\Lambda_{\mathrm{conv}}\,B_{\Phi}}{\sqrt{m}}.
\label{eq:conv-rad}
\end{equation}
\noindent
\textbf{Interpretation.}
Weight regularization decreases $\Lambda_{\mathrm{conv}}$; patch normalization/standardization decreases $B_{\Phi}$.
Parameter sharing (convolution) reduces the number of free parameters relative to an unshared linear map, and is reflected in the fact that
the bound depends on a \emph{single} filter tensor norm rather than per-position independent weights.

\subsection{Rademacher complexity of a TemporalConvLayer (align + causal conv + residual + ReLU)}
We model the TemporalConvLayer as
\begin{equation}
\mathrm{TCL}(x)
~=~
\rho\big(\mathrm{Align}(x) + \mathrm{CausalConv}(x)\big),
\label{eq:tcl-def}
\end{equation}
where $\rho=\mathrm{ReLU}$ is $1$-Lipschitz.
The align module is a $1\times 1$ convolution (a channel-wise linear map) with weights $W_{\mathrm{align}}$,
and the causal convolution has weights $W_{\mathrm{cc}}$.

Assume the corresponding feature bounds
$\|\Phi_{\mathrm{align}}(x_i)\|_F\le B_{\mathrm{align}}$ and $\|\Phi_{\mathrm{cc}}(x_i)\|_F\le B_{\mathrm{cc}}$,
and norm constraints
$\|W_{\mathrm{align}}\|_F\le \Lambda_{\mathrm{align}}$ and $\|W_{\mathrm{cc}}\|_F\le \Lambda_{\mathrm{cc}}$.
Then by~\eqref{eq:rad-contract} with $L_{\rho}=1$ and~\eqref{eq:rad-sum},
\begin{align}
\hat{\mathfrak{R}}_S(\mathrm{TCL})
&\le
\hat{\mathfrak{R}}_S(\mathrm{Align}) + \hat{\mathfrak{R}}_S(\mathrm{CausalConv})
\label{eq:tcl-sum}\\
&\le
\frac{\Lambda_{\mathrm{align}}\,B_{\mathrm{align}}}{\sqrt{m}}
+
\frac{\Lambda_{\mathrm{cc}}\,B_{\mathrm{cc}}}{\sqrt{m}}.
\label{eq:tcl-bound}
\end{align}
\noindent
\textbf{Remark.}
This bound is \emph{distribution-free} and does not require plugging in numerical norms from initialization.
Using initialization statistics (e.g., Xavier) produces only an \emph{expected} norm at $t=0$ and is generally not a generalization certificate after training.

\subsection{Network-level bound via Lipschitz composition}
Let the full network be a composition of $L$ blocks,
$f = f_L\circ \cdots \circ f_1$,
and assume each block $f_\ell$ is $L_\ell$-Lipschitz with respect to the input norm used in the complexity analysis.
Then, for any function class $\mathcal{F}_0$ applied to inputs (e.g., the identity class on bounded inputs),
the composition satisfies the standard Lipschitz propagation:
\begin{equation}
\hat{\mathfrak{R}}_S(f\circ \mathcal{F}_0)
~\le~
\Big(\prod_{\ell=1}^{L} L_\ell\Big)\,
\hat{\mathfrak{R}}_S(\mathcal{F}_0).
\label{eq:lip-prop}
\end{equation}
When each block is a sum of linear convolutions and $1$-Lipschitz activations (as in~\eqref{eq:tcl-def}),
a conservative bound is obtained by taking
\begin{equation}
L_\ell
~\le~
L_{\rho}\Big(L_{\mathrm{align},\ell}+L_{\mathrm{cc},\ell}\Big)
~=~
L_{\mathrm{align},\ell}+L_{\mathrm{cc},\ell},
\end{equation}
where $L_{\mathrm{align},\ell}$ and $L_{\mathrm{cc},\ell}$ can be upper-bounded by the corresponding operator norms
(e.g., spectral norms) of the convolutional linear maps.
Combining~\eqref{eq:lip-prop} with the generalization bound~\eqref{eq:gen-bound}
shows that controlling block Lipschitz constants (e.g., via spectral/weight norm regularization) and keeping feature magnitudes bounded
reduces the complexity term and thus improves generalization.

\subsection{Discussion: parameter sharing and regularization}
Convolutional parameter sharing reduces degrees of freedom relative to fully-connected layers on the flattened spatiotemporal input.
In the above bounds, this effect is reflected by the dependence on \emph{filter norms} ($\Lambda_{\mathrm{conv}}$) rather than on per-location parameters.
Weight decay (or explicit norm constraints) directly reduces $\Lambda_{\mathrm{align}}$ and $\Lambda_{\mathrm{cc}}$,
while normalization/standardization reduces $B_{\mathrm{align}}$ and $B_{\mathrm{cc}}$.
Both mechanisms shrink $\hat{\mathfrak{R}}_S(\cdot)$ at the layer and block level, tightening~\eqref{eq:gen-bound}.

\end{document}